\makeatletter
\def\@fnsymbol#1{\ensuremath{\ifcase#1\or \dagger\or \ddagger\or
   \mathsection\or \mathparagraph\or \|\or **\or \dagger\dagger
   \or \ddagger\ddagger \else\@ctrerr\fi}}
\documentclass[10pt,twocolumn,letterpaper]{article}

\usepackage[pagenumbers]{cvpr} 

\definecolor{cvprblue}{rgb}{0.21,0.49,0.74}
\usepackage{bbold,bbding,colortbl}
\usepackage[pagebackref,breaklinks,colorlinks,allcolors=cvprblue]{hyperref}

\def\paperID{3155} 
\def\confName{CVPR}
\def\confYear{2025}

\title{Free on the Fly: Enhancing Flexibility in Test-Time Adaptation with Online EM}

\author{
Qiyuan Dai \quad Sibei Yang\thanks{Corresponding author}\\
{School of Information Science and Technology, ShanghaiTech University}\\
{Sun Yat-sen University}\\
}

\begin{document}
\maketitle
\begin{abstract}

Vision-Language Models (VLMs) have become prominent in open-world image recognition for their strong generalization abilities. Yet, their effectiveness in practical applications is compromised by domain shifts and distributional changes, especially when test data distributions diverge from training data. Therefore, the paradigm of test-time adaptation (TTA) has emerged, enabling the use of online off-the-shelf data at test time, supporting independent sample predictions, and eliminating reliance on test annotations. 
Traditional TTA methods, however, often rely on costly training or optimization processes, or make unrealistic assumptions about accessing or storing historical training and test data.
Instead, this study proposes FreeTTA, a training-free and universally available method that makes no assumptions, to enhance the flexibility of TTA. 
More importantly, FreeTTA is the first to explicitly model the test data distribution, enabling the use of intrinsic relationships among test samples to enhance predictions of individual samples without simultaneous access—a direction not previously explored. 
FreeTTA achieves these advantages by introducing an online EM algorithm that utilizes zero-shot predictions from VLMs as priors to iteratively compute the posterior probabilities of each online test sample and update parameters. 
Experiments demonstrate that FreeTTA achieves stable and significant improvements compared to state-of-the-art methods across 15 datasets in both cross-domain and out-of-distribution settings.  
\end{abstract}    
\section{Introduction}
\label{sec:intro}
\begin{figure}[t]
    \vspace{0.1cm}
    \centering
    \includegraphics[width=1.05\linewidth]{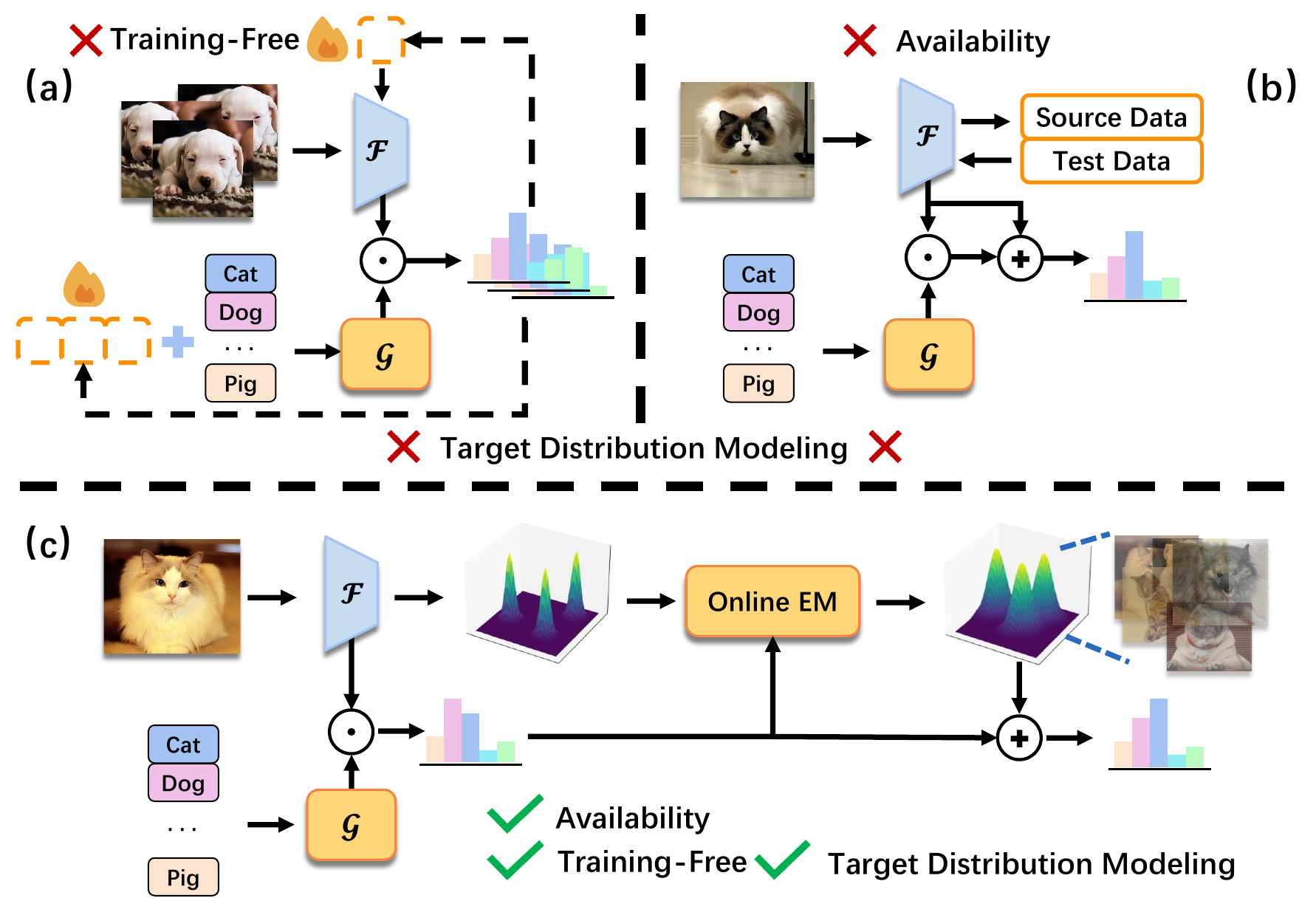}
    \caption{\textbf{Comparison of key properties}: target distribution modeling, availability, and training-free efficiency. (a) Prompt-based methods fail the training-free criterion due to lengthy backpropagation. (b) Other methods require access to source or additional test data, affecting availability. In contrast, (c) our FreeTTA satisfies all criteria, offering a universally available, training-free solution that efficiently models the target distribution without requiring additional assumptions.} 
    \vspace{-6mm}
    \label{fig:intro}
\end{figure}

Vision-Language models (VLMs)~\cite{radford2021learning, jia2021scaling, li2022blip, alayrac2022flamingo}, such as CLIP~\cite{radford2021learning}, pretrained on web-scale image-text pairs, encode a diverse array of visual concepts in a shared embedding space for both image and text.
This enables their zero-shot generalization across various downstream tasks, including image recognition~\cite{zhou2022learning, zhou2022conditional,zhang2024dual,zhang2022tip,khattak2023maple}. Images can be directly identified without task-specific data by aligning their features with the text embeddings of hand-crafted class prompts.
However, despite its strengths, VLMs face notable challenges when downstream images demonstrate distinct domain and distribution shifts relative to their training data in the source domain. 
To address this, some research focuses on adapting VLMs to specific target domains and distributions using parameter-efficient fine-tuning techniques, such as adapter tuning~\cite{zhang2022tip} and prompt learning~\cite{zhou2022learning, zhou2022conditional,zhang2024dual,khattak2023maple}. 
Nonetheless, a key limitation of these works is their assumption of access to sufficient annotated downstream data for fine-tuning, which restricts VLMs' applicability in real-world scenarios where labeled data is unavailable, particularly in dynamic and diverse conditions. 

Given this, the paradigm of utilizing off-the-shelf unlabeled data at test time, known as test-time adaption~\cite{shu2022test,feng2023diverse,karmanov2024efficient,zanella2024test,abdul2024align,sui2024just}, has emerged to improve VLMs' generalization to the target domain. 
Building on prompt learning in VLMs, test-time prompting~\cite{shu2022test} tuning optimizes the prompt by minimizing prediction entropy across test sample augmentations. Follow-up works~\cite{feng2023diverse,karmanov2024efficient,zanella2024test,abdul2024align,sui2024just,yoon2024c} introduce new optimization objectives, such as calibration error~\cite{yoon2024c}, pseudo-label probability difference~\cite{lee2024entropy}, and self-supervised learning metrics~\cite{ma2024swapprompt}, or enhance sample augmentation~\cite{feng2023diverse}. 
More recently, some approaches replace learnable prompts with direct feature representation improvements via meanshift~\cite{zanella2024test} or dynamic adapters~\cite{zhangboostadapter}, or with calibration of the VLMs' temperature factor~\cite{farina2024frustratingly}. 

However, we believe the three key intrinsic characteristics listed below, crucial for fully realizing the core advantages and applications of test-time adaptation in practical settings, are not sufficiently addressed by existing methods. 
\begin{itemize}
    \item \textbf{Target Distribution Modeling.} Explicitly estimating the target distribution can leverage intrinsic relationships among test samples to enhance individual predictions and adapt the model to the overall distribution. Current methods either treat each test sample independently~\cite{shu2022test,feng2023diverse,sui2024just} (Figure~\ref{fig:intro}\textcolor{cvprblue}{a}) or consider only a very small set of sample correlations~\cite{karmanov2024efficient} (Figure~\ref{fig:intro}\textcolor{cvprblue}{b}) limiting test-time adaptation's ability to fully utilize continuously incoming test samples as historical information. 
    \item \textbf{Availability.} Assumptions regarding access to or modification of model parameters, training data, or simultaneous access to or storage of multiple test datasets should be avoided, especially in light of current API access for foundational models and associated privacy concerns. However, prompt tuning methods modify model parameters, while training-free dynamic adapters require storing a subset of test samples~\cite{zhangboostadapter} (Figure~\ref{fig:intro}\textcolor{cvprblue}{b}). 
    \item \textbf{Training-Free for Efficiency and Stability.} There should be no significant increase in inference costs or prediction instability arising from the randomness of optimization and inherent biases in the optimization objective. Unfortunately, most methods significantly increase time complexity~\cite{shu2022test,feng2023diverse,abdul2024align} (Figure~\ref{fig:intro}\textcolor{cvprblue}{a}), while a reduction in entropy is demonstrated to heighten the risk of overfitting and overconfidence. 
\end{itemize}

Therefore, we aim to be the first to enhance test-time adaptation to simultaneously satisfy all three aforementioned characteristics.
As a starting point, we assume that the test samples for each class follow an independent Gaussian distribution, denoted as \( p(x | y = i) \sim N(\mu_i, \Sigma_i) \). This assumption allows us to apply maximum likelihood estimation~\cite{fisher1922mathematical} to determine the distribution parameters (mean \(\mu_i\) and variance \(\Sigma_i\)) for each class, and subsequently classify new samples using Bayes' theorem~\cite{bayes1958essay}, expressed as:
\(
\hat{y} = \arg \max_{i} p(y = i | x) \propto p(x | y = i) p(y = i),
\) 
where \( p(y = i) \) represents the prior probability of class \( i \). 
However, applying Gaussian discriminant analysis~\cite{bishop2006pattern} to classify test samples at test time poses new challenges: (1) The class labels of the test data are unknown, preventing direct estimation of the distribution parameters; and (2) test samples cannot be observed simultaneously, as each is predicted online and sequentially. 

To address challenge (1), we adopt the Gaussian assumption in Gaussian discriminant analysis~\cite{bishop2006pattern} and conclude that the unsupervised test samples conform to a Gaussian mixture distribution, expressed as:
\(
p(x) = \sum_{i=1}^{I} \pi_i \mathcal{N}(x | \mu_i, \Sigma_i),
\)
where \( \pi_i \) represents the mixture weight of class $i$. 
Furthermore, we directly apply the EM algorithm~\cite{moon1996expectation} to iteratively predict the posterior probability \( \gamma_{ki} \) of each sample \( x_k \) belonging to each class \( i \) during the E-step, and update the model parameters \( \mu_i \), \( \Sigma_i \), and \( \pi_i \) based on \( \gamma_{ki} \) and test samples in the M-step. 
To tackle challenge (2), a naive approach involves storing historical test samples during the online prediction process and using them to estimate parameters; however, this contradicts the second characteristic, \ie, availability. 
Instead, we leverage the alignment of visual and textual features in the shared embedding space of the VLM model, using the text embeddings of each class \(i\) as the initial mean \(\mu_i\). We then extend the EM algorithm to an online version, where it sequentially estimates the posterior probability \( \gamma_{ki} \) of the currently arriving test sample and updates the Gaussian mixture parameters in an online manner. 
Additionally, leveraging the generalization capabilities of the VLM's original zero-shot branch, we use its prediction as the confidence score to weigh the influence of each incoming test sample during the online EM update process. 

To evaluate the effectiveness of our FreeTTA and learning strategy, we conduct experiments on cross-domain benchmarks and out-of-distribution benchmarks. In summary, our main contributions are: 
\begin{itemize}
\setlength{\itemsep}{0pt}
\setlength{\parsep}{0pt}
\setlength{\parskip}{0pt}
\item 
To the best of our knowledge, we are the first to simultaneously satisfy the three intrinsic characteristics—target distribution modeling, availability, and being training-free—which enable effective and efficient test-time adaptation for VLMs. 
\item 
We introduce FreeTTA, an online EM method that iteratively predicts posterior probabilities for incoming samples across classes and updates parameters by leveraging the prior knowledge of VLMs. This approach enhances stability and incorporates uncertainty, achieving continuous online adaptation without the need to access or store past or additional data. 
\item 
Experimental results show that our FreeTTA achieves stable and significant improvements compared to state-of-the-art methods across 15 datasets in cross-domain and out-of-distribution settings, highlighting its robustness and effectiveness. 
\end{itemize}

\section{Related Work}

\noindent\textbf{Vision-Language Model.}
In the development of Vision-Language Models (VLMs), several landmark models~\cite{radford2021learning, jia2021scaling, li2022blip, alayrac2022flamingo} continuously advance the boundaries of cross-modal understanding between images and text. Among them, CLIP~\cite{radford2021learning} leverages contrastive learning on large-scale image-text paired data to obtain cross-modal feature representations for both vision and language.
In downstream tasks, these pre-trained VLMs exhibit zero-shot and few-shot learning capabilities, along with advanced semantic understanding, facilitating their broad application across diverse areas. For example, in open-world object detection~\cite{zhong2022regionclip, wu2023cora, kuo2022f, gu2021open}, zero-shot capabilities are extended to detection tasks primarily through knowledge distillation techniques; in image classification~\cite{zhou2022learning, zhou2022conditional}, prompt-based few-shot adaptation is commonly employed to address domain shift;  and in referring image segmentation~\cite{dai2024curriculum, wang2022cris, xu2023bridging, yang2021bottom}, VLMs' advanced semantic alignment facilitate cross-modal alignment at the pixel level, enabling precise segmentation of specified objects within an image.
However, using VLMs in downstream tasks often requires labeled training data to bridge domain gaps.  
Unlike them, our focus is on test-time adaptation, which utilizes off-the-shelf unlabeled data at test time to adapt to the test domain. 

\noindent\textbf{Test-Time Adaptation (TTA).} 
TTA aims to automatically adapt to new data domains or distributions during test time, enabling models to adjust to downstream tasks without requiring additional labeled data. 
This low-cost transfer characteristic holds practical significance and achieves success across various tasks, including image segmentation\cite{chen2024each, wang2023dynamically, yeo2023rapid}, object detection~\cite{veksler2023test}, action recognition~\cite{lin2023video}, and image classification~\cite{shu2022test,feng2023diverse,karmanov2024efficient,zanella2024test,abdul2024align,wang2020tent,zhang2022memo,sui2024just}. In image classification, early TTA approaches utilize classifiers trained solely on the image modality, such as TENT~\cite{wang2020tent},
which first proposes enhancing model confidence by minimizing prediction entropy, thereby generalizing the model to the downstream domain.
With the advancement of VLMs, recent TTA works~\cite{shu2022test,feng2023diverse,karmanov2024efficient,zanella2024test,abdul2024align,sui2024just} adopt CLIP's text encoder as a classifier to leverage its generalization capabilities. Some methods integrate entropy minimization with prompt learning techniques commonly used in CLIP-based classification tasks~\cite{zhou2022learning,zhou2022conditional}. For instance, TPT~\cite{shu2022test} combines entropy minimization with prompt learning, where CLIP is fully frozen and optimizes the learnable text prompt for each test sample. Building on this, DiffTPT~\cite{feng2023diverse} employs a diffusion model to generate additional augmented samples, enhancing robustness. However, these approaches require gradient backpropagation for each test sample to update learnable prompts, leading to high computational costs. 
To enhance efficiency, 
TDA~\cite{karmanov2024efficient} caches historical test data to provide additional pseudo-priors for improving test-time accuracy, circumventing the need for parameter updates. However, these methods generally overlook the potential relationships among test data and often rely on backpropagation or access to additional data, limiting their practicality. In contrast, our approach models the target domain online, featuring both availability and a training-free design.

\noindent\textbf{{GMM \& EM in Computer Vision.}}
In computer vision, Gaussian Mixture Models (GMM)~\cite{reynolds2009gaussian} and the Expectation-Maximization (EM)~\cite{moon1996expectation，cappe2009line} algorithm are widely applied in image classification~\cite{shi2019probabilistic,weber2000unsupervised}, object detection~\cite{choi2019gaussian}, image segmentation~\cite{zhang2001segmentation}, and image generation~\cite{liu2020gmm}. For instance, in image classification, \cite{shi2019probabilistic} proposes a GMM-based approach for modeling facial embeddings, representing them as probability distributions to enhance robustness and accuracy in face recognition. In object detection, \cite{choi2019gaussian} integrates Gaussian distributions into the YOLOv3 model, modeling localization uncertainty to improve detection precision and speed. We introduce the concept of modeling the target domain with GMM into the field of TTA, leveraging an online EM approach to achieve a training-free method without the need for access to additional data.

\section{Method}
\begin{figure*}[th]
\centering
\includegraphics[width=1.05\textwidth]{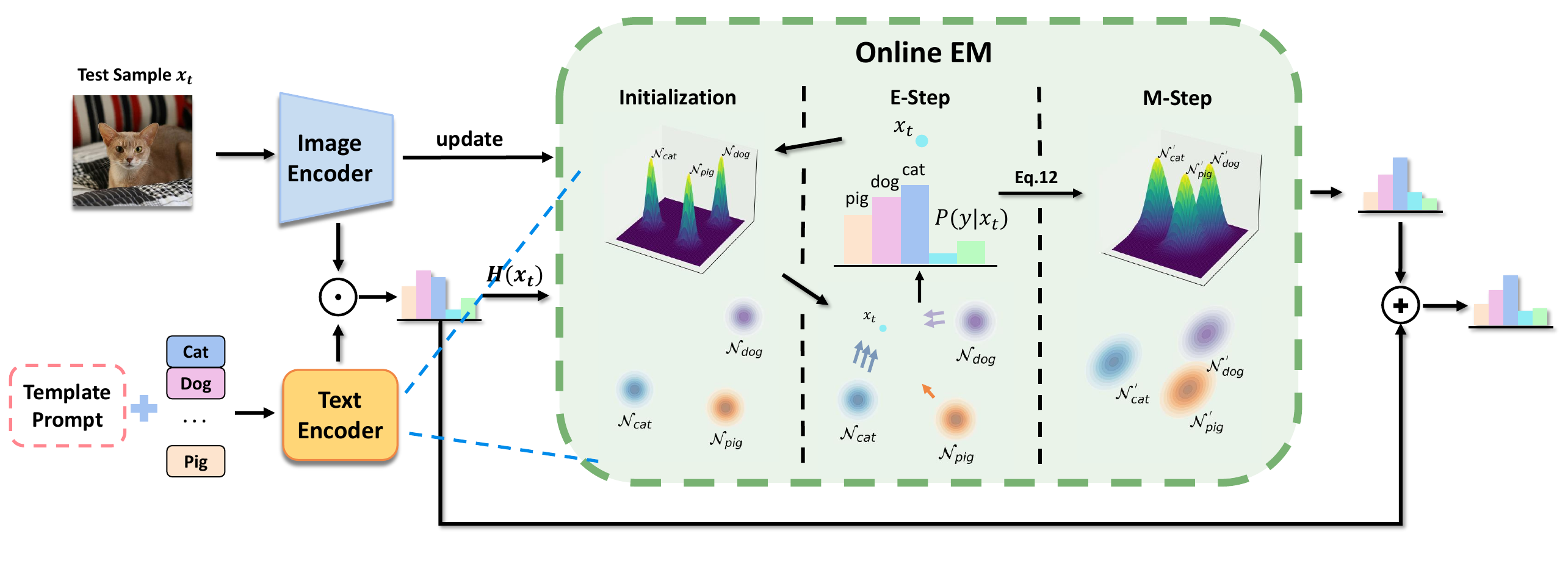}
\vspace{-6mm}
\caption{\textbf{The overall framework of our FreeTTA}. Given a test sample $x_t$, we use the frozen CLIP image encoder to extract the image feature, while the text encoder, using prompt templates, generates class feature vectors. The online EM algorithm is initialized with the text features as mean vectors and the identity matrix as the shared covariance matrix. It updates in two steps: the E-step calculates the posterior probability for each class, and the M-step updates the class-specific mean vectors and shared covariance matrix based on current prediction, leveraging CLIP priors to assess the contribution of each sample. Our FreeTTA combines CLIP’s zero-shot classification results with GDA predictions to enhance stability and robustness in the target domain, explicitly modeling the target distribution without requiring time-consuming training, while meeting the availability requirement.}
\label{fig:frameworks}
\vspace{-0.3cm}
\end{figure*}

\textcolor{black}{The framework of our proposed FreeTTA is shown in Figure~\ref{fig:frameworks}. First, we review the zero-shot CLIP and previous TTA methods for CLIP, along with the challenges they encounter (see Sec~\ref{subsec:preliminary}). Next, we introduce Gaussian discriminant analysis and describe how it models data distributions, followed by a discussion of the challenges in directly applying it to TTA (see Sec~\ref{subsec:gda}). Finally, we provide a detailed introduction of our proposed FreeTTA, demonstrating how it effectively addresses these challenges (see Sec~\ref{subsec:framework}).}

\subsection{Preliminaries}
\label{subsec:preliminary}

\textbf{Vision-Language Models (VLMs)~\cite{radford2021learning,jia2021scaling,alayrac2022flamingo}} like CLIP~\cite{radford2021learning} have recently shown remarkable zero-shot classification capabilities by aligning visual and linguistic modalities. Specifically, CLIP predicts the probability of an image $x$ belonging to class $i$ as follows:
\begin{equation}
P_{\text{CLIP}}(y = i \mid x) = \frac{\exp\left( \cos\left( f(x),\, g(t_i) \right) \right)}{\sum_{k=1}^K \exp\left( \cos\left( f(x),\, g(t_{k}) \right) \right)},
\end{equation}
where $f(\cdot)$ and $g(\cdot)$ are the image and text encoders, respectively, and $\cos(\cdot,\cdot)$ denotes the cosine similarity between features. Here, \(t_i\) is a class-specific description formulated using template prompts for the \(i\)-th class, and \(K\) denotes the number of target categories. 
Despite its strengths, CLIP suffers from performance degradation under domain shifts between the training and test sets. Traditional works~\cite{zhou2022learning,zhou2022conditional,gao2024clip,zhang2022tip} to mitigate domain shift require fine-tuning the model with labeled data from the target domain, which incurs additional labeling and training costs, raising the application barrier for pre-trained VLMs in real-world scenarios. To address this, recent works~\cite{shu2022test,feng2023diverse,karmanov2024efficient,abdul2024align,yoon2024c} introduce test-time adaptation techniques to facilitate VLM adaptation to the target domain without additional labeled data. 

\noindent\textbf{Test-Time Adaption (TTA) for VLMs.} In the TTA setting, a VLM pre-trained on the source domain is adapted to the target domain using only the unlabeled test set $D_{\text{test}} = \{ x_t \}$. This adaptation occurs either through fine-tuning the model in an unsupervised manner or by utilizing a memory of past test samples. 
Notably, each sample is predicted independently during this process. 
To enhance CLIP's classification performance on the target domain, typical TTA methods~\cite{shu2022test,feng2023diverse,abdul2024align} use prompt learning to minimize prediction entropy across multiple augmentations of each test sample. 
This process can be formulated as follows: 
\begin{equation}
\small
    P^*(y \mid x_t) = \frac{1}{\rho M} \sum_{m=1}^{M} \mathbb{1}\left[ \mathcal{H} \left( P_{\text{CLIP}} \left( \mathcal{A}_m(x_t) \right) \right) \leq \tau \right] P_{\text{CLIP}} \left(\mathcal{A}_m(x_{t}) \right),
\end{equation}
where $\mathcal{A}_m(x_t)$ denotes the $m$-th augmented view of $t$-th image $x_t$, $M$ represents the number of augmented views, $\rho$ is the proportion of high-quality augmented views selected, $\tau$ is the self-entropy threshold, and $\mathcal{H}(p) = -\sum_{k=1}^{K} P(y = k \mid x_t) \log P(y = k \mid x_t)$ denotes the self-entropy of the predicted probability distribution over $K$ categories. The optimization objective is to minimize $\mathcal{H}(P^*(y \mid x_t))$. Additionally, other methods~\cite{karmanov2024efficient} cache high-confidence test features to provide reference information for subsequent samples, mitigating domain shift. 

However, existing TTA works face several challenges: (1) they fail to model the target distribution, ignoring intrinsic relationships among test samples; (2) they require access to data beyond the current sample, conflicting with availability in practical settings; and (3) they involve time-consuming optimization, risking instability. To address these, we propose FreeTTA to overcome these limitations. 

\subsection{Gaussian Discriminant Analysis for TTA}
\label{subsec:gda}
We assume that test samples for each class follow an independent Gaussian distribution and employ Gaussian Discriminant Analysis (GDA)~\cite{bishop2006pattern} for classification. Maximum likelihood estimation~\cite{fisher1922mathematical} is used to determine the mean and variance for each class, followed by classifying new samples using Bayes' theorem~\cite{bayes1958essay}. In this section, we first introduce GDA (Sec~\ref{subsubsec:gda}) and then discuss its challenges in applying to TTA (Sec~\ref{subsubsec:challenge}). 

\subsubsection{Gaussian Discriminant Analysis}
\label{subsubsec:gda}
Assume that samples of each class \(y\) follow a multivariate normal distribution, with distinct means and covariances for each class. To simplify, according to \cite{bishop2006pattern}, we assume that all classes share the same covariance matrix $\Sigma$, while the mean vector $\mu_y$ differs for each class. The conditional probability density function for a sample $x_t$ from class $y$ is given by: 
\begin{align}
&p(x_t \mid y) = \mathcal{N}(x_t \mid \mu_y, \Sigma) \nonumber \\
&= \frac{1}{(2\pi)^{d/2} |\Sigma|^{1/2}} \exp\left( -\frac{(x_t - \mu_y)^\top \Sigma^{-1} (x_t - \mu_y)}{2} \right),
\label{cdf}
\end{align}
where $d$ is the dimension of the sample $x_t$, $\mu_y$ is the mean vector for class $y$, and $\Sigma$ is the shared covariance matrix. Eq.~\ref{cdf} represents the Mahalanobis distance between the sample $x_t$ and the class mean $\mu_y$, scaled by the covariance matrix $\Sigma$. This scaling accounts for feature correlations and variance differences across dimensions, yielding a more accurate distribution estimate. 
Based on this assumption, the posterior probability of $y$ can be expressed using Bayes' theorem~\cite{bayes1958essay} as follows:
\begin{equation}
P(y \mid x_t) = \frac{P(y) p(x_t \mid y)}{\sum_{y'} P(y') p(x_t \mid y')},
\end{equation}
where $P(y)$ represents the prior probability of class $y$. Substituting Eq.~\ref{cdf} into the expression yields:
\begin{equation}
P(y \mid x_t) = \frac{\exp\left( -\frac{1}{2} (x_t - \mu_y)^\top \Sigma^{-1} (x_t - \mu_y) \right)}{\sum_{y'} \exp\left( -\frac{1}{2} (x_t - \mu_{y'})^\top \Sigma^{-1} (x_t - \mu_{y'}) \right)}.
\end{equation}
Therefore, for the $t$-th test sample $x_t$, the class with the highest posterior probability is selected as the prediction:
\begin{equation}
y^* = \arg\max_y \left( \log P(y) - \frac{1}{2} (x_t - \mu_y)^\top \Sigma^{-1} (x_t - \mu_y) \right).
\end{equation}

\subsubsection{The Key Challenges of Applying GDA to TTA}
\label{subsubsec:challenge}
Although GDA models the target distribution, it requires a large set of annotated samples for accurate parameter estimates. In TTA, however, test samples are unlabeled and presented sequentially, making direct application of GDA challenging. 

\noindent\textbf{Sequential Online Adaptation.}
In TTA, the model adapts in real-time based solely on incoming test samples, without access to batch data or full distribution. This ``sequential online" nature makes distribution estimation from a single sample unreliable and impossible, especially with substantial shifts. Thus, methods must balance efficiency and accuracy in online updates for effective adaptation. 

\noindent\textbf{Unsupervised Adaptation.}
In TTA, where test sample labels are unavailable, the model must adapt unsupervised. However, GDA relies on labeled samples to estimate parameters. To suit the unsupervised setting, we use a Gaussian mixture model (GMM)~\cite{reynolds2009gaussian} for TTA.

\noindent\textbf{Uncertainty Modeling.} 
Fortunately, we can leverage the zero-shot predictions from VLMs as priors for GDA, serving as pseudo-labels. However, these pseudo-labels carry uncertainty, and the confidence in their predictions must be accounted for. 

\subsection{Online EM Algorithm for TTA}
\label{subsec:framework}
To address the three challenges of applying GDA to TTA, we introduce an online Expectation-Maximization (EM) algorithm that leverages zero-shot predictions from VLMs as priors to iteratively compute the posterior probabilities of each online test sample and update the parameters. 

Specifically, in the E-step (Sec~\ref{subsubsec:E}), we evaluate the posterior probability of the incoming test sample based on the distribution parameters from the previous step. In the M-step (Sec~\ref{subsubsec:M}), we update the mean of each class and the shared covariance to estimate the dynamic changes in the distribution by using the test sample.

\subsubsection{Parameter Initialization}

During initialization, we use CLIP's text encoder \( g(\cdot) \) to generate the class features \( g(t_y) \) for each class \( y \), which serve as the initial mean vectors \( \mu_y \) for that class. We assume that the features are independent and identically distributed with unit variance, setting the shared covariance matrix \( \Sigma \) to the identity matrix \( I \) for a simple and unbiased starting point. The initialization formulas are:
\begin{equation}
\mu_y = g(t_y), \quad \Sigma = I.
\end{equation}
\subsubsection{E-Step: Computing Posterior Probabilities}
\label{subsubsec:E}
In the E-step, for each new test sample $x_t$ at time step $t$, we evaluate the likelihood of $x_t$ belonging to class $y$ by calculating its posterior probability. Based on the assumption of the normal distribution, the posterior probability $P(z_y = 1 \mid x_t)$ is given by:
\begin{equation}
P_{\text{GAUS}}(z_y = 1 \mid x_t) = \frac{\pi_y \cdot \mathcal{N}(x_t \mid \mu_y, \Sigma)}{\sum_{j} \pi_j \cdot \mathcal{N}(x_t \mid \mu_j, \Sigma)},
\end{equation}
where $z_y$ is the indicator variable representing sample $x_t$ belonging to class $y$, and $\pi_y$ denotes the prior probability of class $y$. We denote the posterior probability as $\gamma_{y,t} = P(z_y = 1 \mid x_t)$.

\subsubsection{M-Step: Parameters Update}
\label{subsubsec:M}
In the M-step, we utilize the posterior probabilities $\gamma_{y,t}$ calculated in the E-step to update the parameters for each class, including the prior probability $\pi_y$, the mean vector $\mu_y$, and the shared covariance matrix $\Sigma$.
For each class $y$, the prior probability is updated as:
\begin{equation}
\pi_y' = \frac{N_y + \gamma_{y,t}}{n_t},
\label{pi}
\end{equation}
where \( n_t \) is the total number of samples up to the \( t \)-th step, and \( N_y \) is the number of samples in class \( y \), which is initialized to \(1/\text{number of classes}\).
Meanwhile, the updates to the mean vectors and the shared covariance matrix incorporate the contribution of the new sample to each class, and are formulated as follows: 
\begin{gather}
\mu_y' = \frac{N_y \cdot \mu_y + \gamma_{y,t} \cdot x_t}{N_y + \gamma_{y,t}}, \notag \\ 
\Sigma' = \frac{(n_t - 1)\Sigma + \sum_y \gamma_{y,t} (x_t - \mu_y')(x_t - \mu_y')^\top}{n_t-1}.
\label{musigma}
\end{gather}
Notably, in statistical estimation, the unbiased estimation of the covariance matrix requires dividing by $n_t-1$, accounting for the degrees of freedom lost due to estimating the mean. 
By dynamically adjusting the shared covariance matrix, the model can estimate the distribution of each class, thereby enhancing classification ability on the target domain. 
After incorporating the new sample $x_t$, the sample count for class $y$ is correspondingly updated as $N_y' = N_y + \gamma_{y,t}$.

Our online EM algorithm effectively addresses the distribution shift challenges in TTA for VLMs. By dynamically adjusting class-specific mean vectors, the shared covariance matrix, and prior probabilities, the model adapts to the target domain in an unsupervised, sequential online manner. Moreover, this approach eliminates the need for gradient-based optimization on individual samples, reducing computational overhead and enhancing model robustness. 

\subsection{Incorporating VLM Priors}

During the initial stage of TTA, the model’s instability leads to erratic predictions and updates, causing drift from the original semantic information. To enhance stability, we propose an adaptive update strategy that incorporates VLM priors, including zero-shot classification predictions and confidence levels. Specifically, we use the entropy of CLIP's predictions to assess each sample's confidence level and dynamically adjust its influence on parameter updates, while leveraging the classification logits to adjust the predicted probabilities. 

For the $t$-th test sample $x_t$, we compute the self-entropy based on the CLIP's zero-shot predicted probabilities, denoted as $\{ P_{\text{CLIP}}(z_y = 1 \mid x_t) \mid y = 1, \dots, K \}$, then the self-entropy is defined as:
\begin{equation}
H(x_t) = -\sum_{y=1}^{K} P_{\text{CLIP}}(z_y = 1 \mid x_t) \log P_{\text{CLIP}}(z_y = 1 \mid x_t).
\end{equation}
By incorporating $H(x_t)$ into the previous online EM update steps, we enable the model to adjust the influence of the single sample based on its confidence level. Specifically, we introduce the weighting function $w(h)=e^{-\beta h}$ to revise the previous Eq. ~\ref{pi} and Eq.~\ref{musigma} as follows:
\begin{gather}
\pi_y' = \frac{N_y + w(H(x_t)) \cdot \gamma_{y,t}}{t'_t +w(H(x_t))}, \notag \\
\mu_y' = \frac{N_y \cdot \mu_y + w(H(x_t)) \cdot \gamma_{y,t} \cdot x_t}{N_y + w(H(x_t)) \cdot \gamma_{y,t}}, \notag \\
\textcolor{black}{\Sigma' = \frac{(n_t' - 1) \Sigma + w(H(x_t)) \sum_{y} \gamma_{y,t} (x_t - \mu_y')(x_t - \mu_y')^\top}{n_t' - 1}.}
\end{gather}
The sample count for class $y$ is correspondingly updated as $N_y' = N_y + w(H(x_t)) \cdot \gamma_{y,t}$, \textcolor{black}{and the total number of samples considering uncertainty is updated as ${n_t}'={n_{t-1}}'+ w(H(x_t))$}. This integration of confidence level into the model can effectively reduce the impact of new samples with high uncertainty on the parameter estimation, thereby mitigating the noise issue. Our FreeTTA modulates the contribution of samples based on the confidence of their predictions and adjusts their influence according to self-entropy. This approach emphasizes high-confidence samples, enhancing their dynamic impact on adaptation.

The final predicted logits are a combination of the CLIP's zero-shot logits and those derived from our probabilistic generative model, expressed as:
\begin{equation}
\text{logits}_y = F T_y^\top + \alpha(w_y^\top F + b_y),
\end{equation}
where $F = f(x_t)$ is the image feature, $T_y = g(t_y)$ is the text feature for class $y$, $\alpha$ is a hyper-parameter and $w_y$ and $b_y$ are the weight vector and bias term derived from the probabilistic generative model, with $w_y = \Sigma^{-1} \mu_y$ and $b_y = \log P(y) - \frac{1}{2} \mu_y^\top \Sigma^{-1} \mu_y$, for $y = 1, \dots, K$.
\section{Experiments}

\renewcommand\arraystretch{0.95}
\begin{table*}[ht]
  \centering
  \resizebox{\linewidth}{!}{
    \begin{tabular}{l*{13}{c}>{\columncolor{gray!15}}c}
      \toprule
      Method &T.D &Avail. &T.F. & AIR & CAL & CAR & DTD & EUR & FLWR & FOOD & PETS & SUN & UCF & AVG \\
      \midrule
      CLIP-RN50 &-&-&-& 16.11 & 87.26 & 55.89 & 40.37 & 25.79 & 62.77 & 74.82 & 82.97 & 60.85 & 59.48 & 56.63 \\
      \midrule
      CoOp~\cite{zhou2022learning} &\XSolidBrush&\XSolidBrush&\XSolidBrush& 15.12 & 86.53 & 55.32 & 37.29 & 26.20 & 61.55 & 75.59 & 87.00 & 58.15 & 59.05 & 56.18 \\
      CoCoOp~\cite{zhou2022conditional} &\XSolidBrush&\XSolidBrush&\XSolidBrush& 14.61 & 87.38 & 56.22 & 38.53 & 28.73 & 65.57 & 76.20 & \textbf{88.39} & 59.61 & 57.10 & 57.23 \\
      \midrule
      TPT~\cite{shu2022test} &\XSolidBrush&\CheckmarkBold&\XSolidBrush& 17.58 & 87.02 & 58.46 & 40.84 & 28.33 & 62.69 & 74.88 & 84.49 & 61.46 & 60.82 & 57.66 \\
      DiffTPT~\cite{feng2023diverse} &\XSolidBrush&\CheckmarkBold&\XSolidBrush& 17.60 & 86.89 & \textbf{60.71} & 40.72 & 41.04 & 63.53 & \textbf{79.21} & 83.40 & 62.72 & 62.67 & 59.85 \\
      \midrule
      \rowcolor{gray!15}
       \textbf{Ours}  &\CheckmarkBold&\CheckmarkBold&\CheckmarkBold& \textbf{17.83} & \textbf{90.12} & 58.01 & \textbf{44.21} & \textbf{43.64} & \textbf{68.26} & 77.98 & 86.44 & \textbf{62.84} & \textbf{63.97}  & \textbf{61.33}\\
      \midrule
      \midrule
      CLIP-ViT-B/16 &-&-&-& 23.22 & 93.55 & 66.11 & 45.04 & 50.42 & 66.99 & 82.86 & 86.92 & 65.63 & 65.16 & 64.59 \\
      \midrule
      CoOp~\cite{zhou2022learning} &\XSolidBrush&\XSolidBrush&\XSolidBrush& 18.47 & 93.70 & 64.51 & 41.92 & 46.39 & 68.71 & 85.30 & 89.14 & 64.15 & 66.55 & 63.88 \\
      CoCoOp~\cite{zhou2022conditional} &\XSolidBrush&\XSolidBrush&\XSolidBrush& 22.29 & 93.79 & 64.90 & 45.45 & 39.23 & 70.85 & 83.97 & 90.46 & 66.89 & 68.44 & 64.63 \\
      \midrule
      PromptAlign~\cite{abdul2024align} &\XSolidBrush&\XSolidBrush&\XSolidBrush& 24.80 & 94.01 & 68.50 & 47.24 & 47.86 & 72.39 & 86.65 & 90.76 & 67.54 & 69.47 & 66.92 \\
      TDA~\cite{karmanov2024efficient} &\XSolidBrush&\XSolidBrush&\CheckmarkBold& 23.91 & 94.24 & 67.28 & 47.40 & 58.00 & 71.42 & 86.14 & 88.63 & 67.62 & 70.66 & 67.53 \\
      TPT~\cite{shu2022test} &\XSolidBrush&\CheckmarkBold&\XSolidBrush& 24.78 & 94.16 & 66.87 & 47.75 & 42.44 & 68.98 & 84.67 & 87.79 & 65.50 & 68.04 & 65.10 \\
      DiffTPT~\cite{feng2023diverse} &\XSolidBrush&\CheckmarkBold&\XSolidBrush & 25.60 & 92.49 & 67.01 & 47.00 & 43.13 & 70.10 & 87.23 & 88.22 & 65.74 & 62.67 &65.47 \\
      \midrule
        MTA~\cite{zanella2024test}  &\XSolidBrush&\CheckmarkBold&\CheckmarkBold& \textbf{25.32} & 94.13 & \textbf{68.05} & 45.59 & 38.71 & 68.26 & 84.95 & 88.22 & 64.98 & 68.11  & 64.63\\
         ZERO~\cite{farina2024frustratingly}  &\XSolidBrush&\CheckmarkBold&\CheckmarkBold& 24.40 & 93.51 & 67.54 & 45.80 & 39.60 & 67.07 & 84.36 & 86.74 & 64.49 & 67.64  & 64.66\\
         \rowcolor{gray!15}
        \textbf{Ours} &\CheckmarkBold&\CheckmarkBold&\CheckmarkBold& 25.11 & \textbf{94.63} & 67.34 & \textbf{46.96} & \textbf{62.93} & \textbf{71.62} & \textbf{86.62} & \textbf{90.11} & \textbf{67.76} & \textbf{71.16} & \textbf{68.42}  \\
      \bottomrule
    \end{tabular}
  } 
  \caption{\textbf{Comparison on Cross-domain Benchmark.} The best performance for each dataset is highlighted in \textbf{bold}. Methods are categorized based on three key attributes: target distribution modeling (T.D.), availability (Avail.), and training-free (T.F.) characteristics.}
  \label{tab:fine-grained}
  \vspace{-8pt}
\end{table*}

  
\subsection{Datasets and Implementation Details}
\noindent\textbf{Datasets.} 
To evaluate our method, we first conduct extensive cross-domain generalization experiments across 10 diverse datasets, encompassing image classification tasks across different domains: FGVCAircraft~\cite{maji2013fine}, Caltech101~\cite{fei2004learning}, StanfordCars~\cite{krause20133d}, DTD~\cite{cimpoi2014describing}, EuroSAT~\cite{helber2019eurosat}, Flower102~\cite{nilsback2008automated}, Food101~\cite{bossard2014food}, OxfordPets~\cite{parkhi2012cats}, SUN397~\cite{xiao2010sun}, and UCF101~\cite{kay2017kinetics}. This selection ensures a comprehensive assessment of the model’s adaptability across varied visual domains.
To further evaluate the robustness of our method under natural distribution shifts, we utilize the ImageNet~\cite{deng2009imagenet} dataset along with its challenging out-of-distribution (OOD) variants: ImageNet-A~\cite{hendrycks2021natural}, ImageNet-V2~\cite{recht2019imagenet}, ImageNet-R~\cite{hendrycks2021many}, and ImageNet-S~\cite{wang2019learning}. These benchmarks are designed to test the model's resilience to natural variations in data distribution.

\noindent\textbf{Implementation Details.} 
We follow prior work by utilizing the pre-trained CLIP model with either ResNet-50 or ViT-B/16 as the image encoder, paired with their respective text encoders. For class labels across different datasets, we follow TDA~\cite{karmanov2024efficient}, employing specific template prompts for each dataset, which are processed by the text encoder to serve as the zero-shot CLIP classifier. For our FreeTTA, we set $\alpha$ to 0.2 and $\beta$ to 4.5. During testing, we strictly adhere to the TTA setting in ~\cite{shu2022test}, using a batch size of 1. We use top-1 accuracy as the evaluation metric. All experiments are conducted on an NVIDIA 3090 GPU.

\subsection{Comparison with the State-of-the-Art Methods}
Table~\ref{tab:fine-grained} and Table~\ref{tab:ood-main} present the comparisons on the cross-domain and out-of-distribution benchmarks, respectively, against state-of-the-art methods. Among them, CoOp~\cite{zhou2022learning} and CoCoOp~\cite{zhou2022conditional} are few-shot adaptation methods that require labeled data from the target domain for prompt learning. 
For test-time adaptation methods, we categorize them based on three key attributes: target distribution modeling, availability, and training-free characteristics. Compared with other methods that also possess availability and training-free characteristics, our approach achieves stable improvements across diverse datasets, with an average gain of 3.76\% on cross-domain benchmark and 1.66\% on out-of-distribution benchmark. 
\renewcommand\arraystretch{0.7}
\begin{table*}[ht]
  \centering
  \small
  \resizebox{0.8\linewidth}{!}{
  \begin{tabular}{l*{10}c}
    \toprule
    {Method}     &T.D &Avail. &T.F.  & ImageNet   &  -A  &  -V2  & -R  & -S   &{Average}  & {OOD Average}     \\
  \midrule
  CLIP-RN50      &-&-&-  &59.81&	23.24&	52.91	&{60.72}	&35.48&	46.43&	43.09           \\ 
  \midrule
  CoOp~\cite{zhou2022learning}   &\XSolidBrush&\XSolidBrush&\XSolidBrush &  63.33 &  23.06  &   55.40     &  56.60   &  34.67   &     46.61   &     42.43       \\
  
  CoCoOp~\cite{zhou2022conditional}    &\XSolidBrush&\XSolidBrush&\XSolidBrush      &  62.81 &  23.32  &   55.72     &  57.74  &  34.48   &     46.81   &    42.82   \\
Tip-Adapter &\XSolidBrush&\XSolidBrush&\CheckmarkBold & 62.03 & 23.13 & 53.97 & 60.35  & 35.74 & 47.04 & 43.30 \\
  \midrule
  TPT     &\XSolidBrush&\CheckmarkBold&\XSolidBrush      &  60.74 & 26.67  &    54.70   &  59.11   &  35.09   &    47.26    &     43.89    \\
  DiffTPT&\XSolidBrush&\CheckmarkBold&\XSolidBrush  & 60.80 & \textbf{31.06} & 55.80 & 58.80 & 37.10 & 48.71 & 45.69 \\
  \rowcolor{gray!15}
    \textbf{Ours} &\CheckmarkBold&\CheckmarkBold&\CheckmarkBold& \textbf{61.51} & 30.67 & \textbf{55.89} & \textbf{63.02} & \textbf{37.94} & \textbf{49.81} & \textbf{46.88} \\
  \midrule
  \midrule
  CLIP-ViT-B/16    &-&-&- & 68.34&	49.89&	61.88&	{77.65}&	48.24&	61.20&	59.42             \\
  \midrule
  CoOp~\cite{zhou2022learning} &\XSolidBrush&\XSolidBrush&\XSolidBrush  &  71.51 &  49.71  &   64.20     &  75.21   &  47.99   &   61.72     &   59.28  \\
  
  CoCoOp~\cite{zhou2022conditional}  &\XSolidBrush&\XSolidBrush&\XSolidBrush  &  71.02  & 50.63  & 64.07       &  76.18   &  48.75   &   62.13     &   59.91  \\
 Tip-Adapter &\XSolidBrush&\XSolidBrush&\CheckmarkBold  & 70.75 & 51.04 & 63.41 & 77.76 & 48.88 & 62.37 & 60.27 \\
  \midrule
  PromptAlign~\cite{abdul2024align}&\XSolidBrush&\XSolidBrush&\XSolidBrush  & - & 59.37 & 65.29 & 79.33 & 50.23 & - & 63.55 \\
    TDA~\cite{karmanov2024efficient} &\XSolidBrush&\XSolidBrush&\CheckmarkBold  & 69.51 & 60.11 & 64.67 & 80.24 & 50.54 & 65.01 & 63.89 \\
  TPT~\cite{shu2022test}   &\XSolidBrush&\CheckmarkBold&\XSolidBrush  &  68.98  & 54.77  & 63.45       &  77.06   &  47.94   &   62.44     &   60.81 \\
  DiffTPT~\cite{feng2023diverse}  &\XSolidBrush&\CheckmarkBold&\XSolidBrush &  70.30  & 55.68  & 65.10       &  75.00   &  46.80   &   62.28     &   60.52 \\
    \midrule
   MTA~\cite{zanella2024test} &\XSolidBrush&\CheckmarkBold&\CheckmarkBold & 69.29 & 57.41 & 63.61 & 76.92 & 48.58 & 63.16 & 61.63 \\
  ZERO~\cite{farina2024frustratingly} &\XSolidBrush&\CheckmarkBold&\CheckmarkBold & 69.06 & 61.35 & 64.13 & 77.28 & 48.29 & 64.02 & 62.76 \\
  \rowcolor{gray!15}
  \textbf{Ours} &\CheckmarkBold&\CheckmarkBold&\CheckmarkBold & \textbf{70.21} & \textbf{61.41} & \textbf{64.92} & \textbf{80.49} & \textbf{50.88} & \textbf{65.58} & \textbf{64.42} \\
    \bottomrule
  \end{tabular}
}
\caption{\textbf{Comparison on OOD Benchmark.} The best performance for each dataset is highlighted in \textbf{bold}. Methods are categorized according to the same criteria as in Table~\ref{tab:fine-grained}. The OOD average reflects the mean performance across the four ImageNet variant datasets.}
\label{tab:ood-main}
\vspace{-5pt}
\end{table*}
  
\noindent\textbf{Comparison on Cross-Domain Benchmark.}
Recent methods, such as MTA~\cite{zanella2024test} and ZERO~\cite{farina2024frustratingly}, address both availability and training-free issues but fail to leverage potential relationships between test samples. 
Compared to them, our FreeTTA consistently leads on 8 out of 10 datasets, showing average accuracy improvements of 3.99\% and 1.58\%, respectively. This demonstrates our design of explicitly modeling the target distribution through online EM enables more effective adaptation to the target domain. Furthermore, even compared to other TTA methods~\cite{abdul2024align,karmanov2024efficient} that are not training-free or lack availability, we still outperform them on most datasets. Our FreeTTA achieves a 1.5\% improvement in average accuracy over PromptAlign~\cite{abdul2024align}, which requires source domain statistics and not being training-free, underscoring the computational efficiency and adaptation capabilities of our approach. Moreover, we outperform TDA~\cite{karmanov2024efficient} on 9 out of 10 datasets with an average improvement of 0.89\%, which TDA necessitates an explicit cache of test sample features and uses them solely as instance-level references. In contrast, our method eliminates the need for such storage and models the target domain using an online EM approach. Compared to TPT~\cite{shu2022test}, the pioneering TTA algorithm in VLMs, our method surpasses it on both ResNet-50 and ViT-B/16, with average improvements of 3.67\% and 3.32\%, respectively, demonstrating the effectiveness of our approach across different backbones.

\noindent\textbf{Comparison on OOD Benchmark.}
Furthermore, we conduct additional comparisons with other methods on OOD datasets that focus on natural distribution shifts as shown in Table~\ref{tab:ood-main}.
Our method outperforms MTA~\cite{zanella2024test} and ZERO~\cite{farina2024frustratingly} across all datasets due to its target distribution modeling capabilities, achieving average accuracy increases of 2.42\% and 1.56\%, and OOD accuracy gains of 2.79\% and 1.66\%. These consistent performance improvements highlight the effectiveness of our method for target distribution modeling while maintaining the advantages of being training-free and not requiring access to the source domain. Additionally, we also achieve average OOD accuracy improvements of 0.87\% and 0.53\%, compared with PromptAlign~\cite{abdul2024align} and TDA~\cite{karmanov2024efficient}. Even greater gains are observed when compared to TPT~\cite{shu2022test} and DiffTPT~\cite{feng2023diverse}, with average improvements of 3.14\% and 3.3\% and OOD accuracy gains of 3.61\% and 3.9\%. 
Our method consistently achieves performance gains, whether compared to traditional prompt-learning-based approaches or recent methods that possess availability or training-free characteristics, which demonstrates the robustness and effectiveness of our target distribution modeling design.

\subsection{Ablation Study}
We conduct ablation studies to demonstrate the effectiveness of our method, as shown in Table~\ref{tab:ablation}, and we use zero-shot CLIP-ViT-B/16  as the baseline (row 1).\\
\noindent\textbf{Update Mean Vectors.}
To verify the importance of dynamically updating the mean vectors $\mu_y$ in our approach, we perform an ablation experiment using fixed mean vectors (row 3). In this experiment, the model utilizes mean vectors initialized with CLIP text embeddings without updates during testing. The results indicate that the model with fixed mean vectors fails to adapt the class feature centers effectively under significant distribution shifts, resulting in decreased classification accuracy. This highlights that the dynamic update of mean vectors is a crucial mechanism in our method, enabling the model to leverage inter-sample relationships in the visual branch and address the generalization challenges of the CLIP on target domains.\\
\noindent\textbf{Update Covariance Matrix.}
We further investigate the necessity of dynamically updating the covariance matrix $\Sigma$ by conducting experiments with a fixed covariance matrix (row 4). In this setup, the covariance matrix is set as an identity matrix and remains unchanged during testing, reducing Equ~\ref{cdf} to a Euclidean distance metric. The results demonstrate that a fixed covariance matrix limits the model's ability to represent intra-class variability, leading to suboptimal performance when adapting to the target domain. In contrast, our method with dynamically updated covariance matrices captures class distribution variations more effectively, thereby improving classification performance at test time.\\
\begin{table}[]
\centering
\small
\renewcommand{\arraystretch}{1.07}
\tabcolsep=0.143cm
\begin{tabular}{l|l|c}
\toprule[1pt]
\multicolumn{1}{c|}{} &Method &\multicolumn{1}{c}{Average Accuracy} \\
\hline
1&Zero-Shot CLIP                           & 64.59       \\ 
2&FreeTTA                     &  68.42   \\
3&2-\text{Mean Vectors Update}                       &  64.64  \\ 
4&2-\text{Covariance Matrix Update}                  &  67.07 \\
5&2-\text{VLM priors}                  &   67.78  \\

\bottomrule[1pt]
\end{tabular}
\caption{\textbf{Ablation Study} on Cross-Domain Benchmark. The CLIP ViT-B/16 is used.}
\label{tab:ablation}
\vspace{-5mm}
\end{table}
\noindent\textbf{VLM priors.}
Finally, we evaluate the impact of incorporating VLM priors for parameter initialization and adjusting influence on parameter updates by assessing each sample's confidence level (row 5). The results show that without consideration of VLM priors, the model's performance decreases due to noise from high-uncertainty samples. Conversely, incorporating self-entropy as an uncertainty measure allows the model to leverage its knowledge as priors to assess the contribution of samples adaptively, making it more robust and stable against noise.
\section{Conclusion}
This paper introduces a novel test-time adaptation approach for VLMs, leveraging the Gaussian discriminant analysis and an adaptive online EM algorithm to improve adaptability under domain shifts. By incorporating VLM priors as uncertainty measurement, our method effectively handles varying sequential online samples and enhances model stability during adaptation. Experimental results demonstrate that our approach significantly improves performance without relying on source domain data and costly training, showcasing its robustness and efficiency.
\textbf{Acknowledgment:} 
This work was supported by the National Natural Science Foundation of China (No.62206174).
{
    \small
    \bibliographystyle{ieeenat_fullname}
    \bibliography{main}
}

\end{document}